\documentclass[final]{cvpr}

\usepackage{times}
\usepackage{epsfig}
\usepackage{graphicx}
\usepackage{amsmath}
\usepackage{amssymb}

\usepackage{amsthm}
\usepackage{float}
\usepackage{mathtools}
\usepackage{xcolor}
\usepackage{float}
\usepackage{url}
\usepackage[utf8]{inputenc}
\usepackage{float}
\usepackage{subcaption}
\usepackage{wrapfig}
\usepackage{lipsum}
\usepackage[ruled,linesnumbered]{algorithm2e}
\usepackage{algorithmic}
\SetKwInput{KwInput}{Input}
\SetKwInput{KwOutput}{Output}

\DeclareMathOperator*{\argmin}{argmin}
\DeclareMathOperator*{\argmax}{argmax}
\newlength{\tempdima}
\newcommand{\rowname}[1]
{\rotatebox{90}{\makebox[\tempdima][c]{\textbf{#1}}}}
\def \RR {\mathbb{R}}

\usepackage[pagebackref=true,breaklinks=true,colorlinks,bookmarks=false]{hyperref}

\def\cvprPaperID{****} 
\def\confYear{CVPR 2021}

\begin{document}

\title{Adversarial Laser Beam: Effective Physical-World Attack to DNNs in a Blink}

\author{Ranjie Duan\textsuperscript{1}\footnotemark[2] \ \  
Xiaofeng Mao\textsuperscript{2}\ \  
A. K. Qin\textsuperscript{1} \ \
Yun Yang\textsuperscript{1} \ \ 
Yuefeng Chen\textsuperscript{2}\ \
Shaokai Ye\textsuperscript{3}\ \
Yuan He\textsuperscript{2}\\
\textsuperscript{1}Swinburne University of Technology \ \ 
\textsuperscript{2}Alibaba Group\\
\textsuperscript{3}EPFL \ \  
}
\maketitle

\renewcommand{\thefootnote}{\fnsymbol{footnote}} 
\footnotetext[2]{Works done when intern at Alibaba} 
\footnotetext[3]{Code is available at https://github.com/RjDuan/Advlight}
\begin{abstract}
Though it is well known that the performance of deep neural networks (DNNs) degrades under certain light conditions, there exists no study on the threats of light beams emitted from some physical source as adversarial attacker on DNNs in a real-world scenario. 
In this work, we show by simply using a laser beam that DNNs are easily fooled.  
To this end, we propose a novel attack method called Adversarial Laser Beam ($AdvLB$), which enables manipulation of laser beam's physical parameters to perform adversarial attack. Experiments demonstrate the effectiveness of our proposed approach in both digital- and physical-settings. We further empirically analyze the evaluation results and reveal that the proposed laser beam attack may lead to some interesting prediction errors of the state-of-the-art DNNs. 
We envisage that the proposed $AdvLB$ method enriches the current family of adversarial attacks and builds the foundation for future robustness studies for light. 
\end{abstract}

\section{Introduction}
\begin{figure}[ht]
\setlength{\abovecaptionskip}{-0.2cm}
 \begin{center}
    \includegraphics[width = \linewidth]{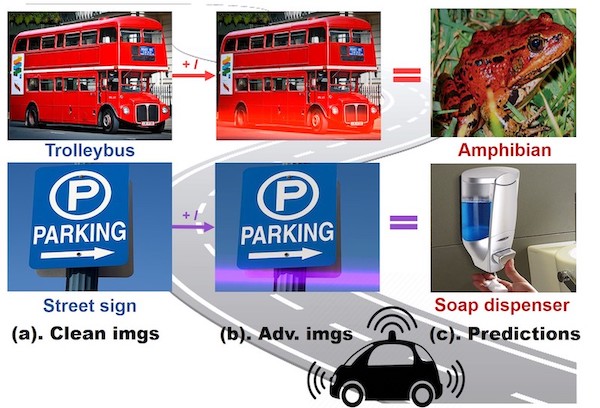}
  \end{center}
  \caption{\textbf{An example.} When the camera of self-driving car captures object shot by the laser beam, it recognizes "trolleybus" as "amphibian" and "street sign" as "soap dispenser". }
  \label{fig:scenario}
\end{figure}

Natural phenomena may play the role of adversarial attackers, \eg a blinding glare results in a fatal crash of a Tesla self-driving car.
What if a beam of light can adversarially attack a DNN? Further, how about using a beam of light, specifically the laser beam, as the weapon to perform attacks. If we can do that, with the fastest speed in the world, the laser beam could achieve the fastest attack with no doubts. 
As shown in Figure \ref{fig:scenario}, by using an off-the-shelf lighting device such as a laser pointer, the attacker can maliciously shoot a laser beam onto the target object to make the self-driving car fail to recognize target objects correctly.

We regard the attack illustrated in Figure \ref{fig:scenario} as a new type of adversarial attack, which is crucial but not yet exploited. Up to now, most researchers study the security of DNNs by exploring various adversarial attacks in digital-settings, where input images are added with deliberately crafted perturbations and then fed to the target DNN model \cite{szegedy2013intriguing,goodfellow2014explaining,dong2018boosting,carlini2017adversarial,madry2017towards}.
However, in physical-world scenarios, images are typically captured by cameras and then directly fed to the target models, where attackers cannot directly manipulate the input image. Some recent efforts in developing physical-world attacks are addressed in \cite{sharif2016accessorize, evtimov2017robust, brown2017adversarial, duan2020adversarial, huang2020universal}. 
The physical-world adversarial examples typically require large perturbations, because small perturbations are hard to be captured by cameras. In addition, the attacking effects of adversarial examples of small perturbations can be easily mitigated in complex physical-world environments \cite{sharif2016accessorize, eykholt2018robust, duan2020adversarial}.
Meanwhile, physical-world adversarial examples require high stealthiness to avoid being discovered by either the victim or defender before performing an attack successfully. Thus for creating physical-world adversarial examples, there is always a compromise between stealthiness and adversarial strength.

Most existing physical-world attacks adopt a "sticker-pasting" setting, \ie, the attacker prints adversarial perturbation as a sticker and then pastes it onto the target object \cite{liu2019perceptual, brown2017adversarial, duan2020adversarial,evtimov2017robust}. These attacks achieve the stealthiness of adversaries with extra efforts of designing adversarial perturbation or camouflaging adversarial images and finding the most effective area in the target object to impose them \cite{liu2019perceptual,xu2019adversarial,sharif2016accessorize,duan2020adversarial}. Besides the challenge of stealthiness, the "sticker-pasting" setting also requires the target object to be physically accessible by the attacker to paste stickers. However, this may not be always possible. Several works explore physical-world threats beyond the "sticker-pasting" setting: Shen et al. \cite{shen2019vla} and Nguyen et al. \cite{nguyen2020adversarial} proposed using a projector to project the adversarial perturbation on the target to perform an attack. However, these works still require manual effort to craft adversarial perturbation.



\begin{figure}[!t]
  \centering
\begin{subfigure}[b]{0.30\linewidth}
    \includegraphics[width = \linewidth]{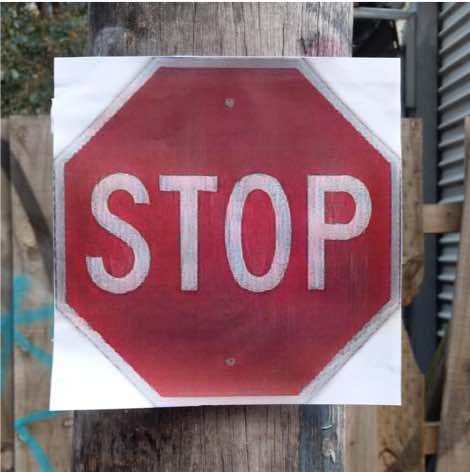}
    \caption{$AdvCam$ \cite{duan2020adversarial}}
  \end{subfigure}
  \begin{subfigure}[b]{0.30\linewidth}
    \includegraphics[width = \linewidth]{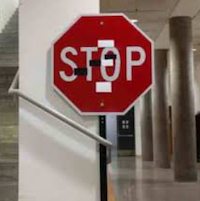}
    \caption{$RP_2$ \cite{brown2017adversarial}}
  \end{subfigure}
  \begin{subfigure}[b]{0.30\linewidth}
    \includegraphics[width = \linewidth]{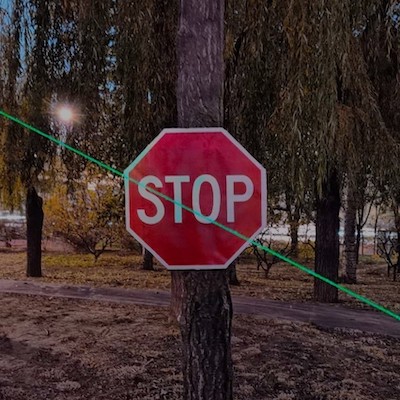}
    \caption{$AdvLB$ (Ours)}
  \end{subfigure}
  \caption{\textbf{Visual comprison.}}
  \label{fig:other_work}
\end{figure}

In our work, we propose a new type of physical-world attack, named adversarial laser beam ($AdvLB$). Unlike existing methods, we utilize the laser beam as adversarial perturbation directly rather than crafting adversarial perturbation from scratch. As $AdvLB$ does not require physically changing the target object to be attacked as in the "sticker-pasting" setting, it features high flexibility to attack any object actively, even from a long distance. In terms of stealthiness, a visual comparison between our proposed attack and other works can be seen in Figure \ref{fig:other_work}. 
Though the adversarial example generated via $AdvLB$ may appear less stealthy than some generated by other approaches such as $AdvCam$. $AdvLB$ may introduce high temporal stealthiness due to its unique physical-attack mechanism. Specifically, with the nature of light, $AdvLB$ can perform the attack in a blink right before the attacked target object gets captured by the imaging sensor, and thus avoid being noticed by victims and defenders in advance. Existing works focus on security issues of DNNs in the daytime whilst potential security threats at night are often ignored. Our proposed $AdvLB$ provides a complementary in this regard. Figure \ref{fig:other_work} illustrates the advantage of $AdvLB$ when the lighting condition is poor.
To launch such an attack, we formulate the laser beam with a group of controllable parameters. Then we devise an optimization method to search for the most effective laser beam's physical parameters in a greedy manner. It enables finding where and how to make an effective physical-world attack in a black-box setting. We further apply a strategy called $k$-random-restart to avoid falling into the local optimum, which increases the attack success rate. 

We conduct extensive experiments for evaluation of the proposed $AdvLB$. We first evaluate $AdvLB$ in a digital-setting, which is able to achieve $95.1\%$ attack success rate on a subset of ImageNet. We further design two physical-world attacks including indoor and outdoor tests, achieving $100\%$ and $77.43\%$ attack success rates respectively. Then ablation studies are presented on the proposed $AdvLB$. Furthermore, we analyze the prediction errors of DNNs caused by the laser beam. We find the causes of prediction errors could be roughly divided into two categories. 1). The laser beam's color feature changes the raw image and forms a new cue for DNNs. 2). The laser beam introduces some dominant features of specific classes, especially lighting related classes, \eg candle. When the laser beam and target object appear simultaneously, there is a chance that the DNN is more biased towards the feature introduced by the laser beam and thus resulting in misclassification. These interesting empirical findings open a door for both attackers and defenders to investigate how to better manipulate this new type of attack. Our major contributions are summarized as follows:
\begin{itemize}
    \item We propose a new type of physical-world attack based on the laser beam named $AdvLB$, which leverages light itself as adversarial perturbation. It provides high flexibility for attacker to perform the attack in a blink. Besides, the deployment of such attack is rather simple: by using a laser pointer, it may become a common threat due to its convenience to perform attack. 
    \item We conduct comprehensive experiments to demonstrate the effectiveness of $AdvLB$. Specifically, we perform physical test to validate $AdvLB$ and show the real-world threats of laser beam by simply using a laser pointer. Thus $AdvLB$ can be a useful tool to explore such threats in the real-world. 
    \item We make an in-depth analysis of the prediction errors caused by the $AdvLB$ attack to have revealed some interesting findings which would motivate future study on $AdvLB$ from the perspectives of attackers and defenders.
\end{itemize}

\section{Background and Related Work}\label{sec:bg}
Adversarial attack was first proposed by Szegedy et al. \cite{szegedy2013intriguing}, aiming to generate perturbations superimposed on clean images to fool a target model. Given a target model $f$, adversarial examples can be crafted by one or more steps of perturbation following the direction of adversarial gradients \cite{goodfellow2014explaining,kurakin2016adversarial, madry2017towards} or optimized perturbation with a given loss \cite{carlini2017adversarial,chen2018ead}. 
Adversarial examples can be either generated from an image itself (in the digital setting) or produced by capturing an attacked physical scene via image sensors such as cameras (in the physical setting) \cite{kurakin2016adversarial}.

\textbf{Adversarial attack in digital settings.}~Most attacks are developed in a digital setting. And their perturbations are bounded by a small norm-ball to ensure that imperceptible to human observers. Normally, $l_{2}$ and $l_{\infty}$ are the most commonly used norms \cite{carlini2017adversarial,carlini2017towards,xie2019improving,dong2018boosting,madry2017towards}. Some other works also explore adversarial examples beyond bounded setting. They make modifications on the secondary attributes (\eg color \cite{hosseini2018semantic,shamsabadi2020colorfool, zhao2020towards}, texture \cite{wiyatno2019physical}) to generate adversarial examples. Besides, there are also several works that propose changing physical parameters \cite{zeng2019adversarial,liu2018beyond} while preserving critical components of images to create adversaries. However, digital attacks have a strong assumption that the attacker has access to modify the input image of the target model directly, which is not practical in real-world scenarios.

\textbf{Adversarial attack in physical settings.}~Kurakin et al. \cite{kurakin2016adversarial} first showed the existence of adversarial examples in the physical-world by printing digital adversary in the physical-world, and then recaptured by cameras. \cite{kurakin2016adversarial}, and its follow-up work adopted such setting, including pasting adversarial patch on either traffic-sign \cite{evtimov2017robust,duan2020adversarial} or t-shirt \cite{xu2019adversarial}, or camouflaging the adversarial patch into specific shape (\eg eye-glasses frames \cite{sharif2016accessorize, brown2017adversarial}, \etc). Due to various physical-world conditions such as viewpoint shifts, camera noise, and other natural transformations \cite{athalye2017synthesizing}, the physically realized adversarial examples always require various adaptations over a distribution of transformations to adapt to physical-world conditions \cite{sharif2016accessorize, evtimov2017robust, brown2017adversarial}. Thus the "sticker-pasting" attacks generate large adversarial perturbation inevitably. However, there exist a certain period of time between deploying the "sticker-pasting" attacks and performing attacks successfully. Thus the "sticker-pasting" attacks require high stealthiness to avoid being noticed by human observers in advance. 
Stealthiness for large perturbation is always a challenge for "sticker-pasting" physical-world attacks \cite{eykholt2018robust, brown2017adversarial,sharif2016accessorize}. Also, these attacks require attacker physically pasting the stickers on the target objects, however, not every object is easily accessible or reachable in real-world, \eg traffic sign on a high pole. 
Crafted perturbation in physical setting suffers from the loss in adversarial strength when converting from digital to physical-setting. In contrary, our method simply leverages the light itself as adversarial perturbation. 

\textbf{Physical attacks with lighting devices.}~There exist some works using lighting devices to generate adversarial attacks. Some utilized a projector to perform the physical-world attacks against face recognition systems \cite{shen2019vla, nguyen2020adversarial}. These attacks craft adversarial perturbation and then project the perturbation onto the target to perform the attack. Zhou et al. \cite{zhou2018invisible} proposed to deploy LED light on the cap to fool the face recognition systems, which requires careful deployment of the lighting device. Comparatively, existing methods require efforts to craft the adversarial perturbation while our method is much easier to perform an adversarial attack.

\section{Approach}\label{sec:approach}
Recall the typical definition of adversarial example $x_{adv}$ in image classification, given an input image $x \in \RR^m$ with class label $y$, a DNN classifier $f: \RR^m \to \RR^J$. The classifier $f$ associates with a confidence score $f_{j}(x)$ to class $j$. An adversarial example $x_{adv}$ commonly satisfies two properties: 1) $\argmax_{j \in J} f(x_{adv}) \neq \argmax_{j \in J}f(x)$, 2) $\left\|x_{adv}-x\right\|_p \leq \epsilon$. In which, the first property requires that $x_{adv}$ is classified differently with $x$ by target model $f$. The second property requires that the perturbation of $x_{adv}$ is imperceptible enough so that $x_{adv}$ is stealthy for human observers.



\begin{figure}[htb]
  \setlength{\abovecaptionskip}{-0.2cm}
 \begin{center}
    \includegraphics[width = \linewidth]{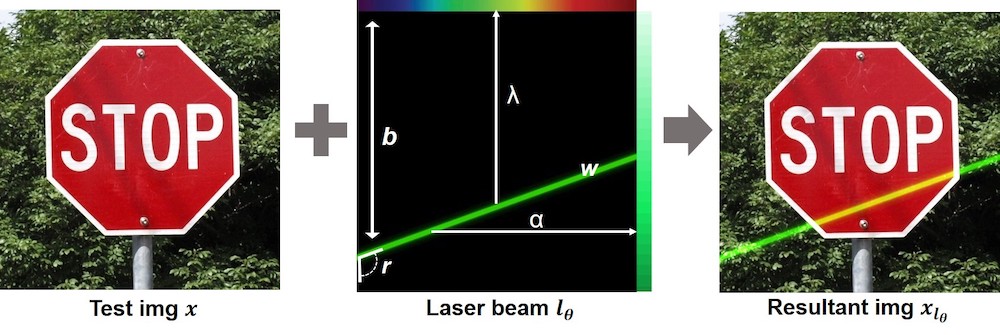}
  \end{center}
  \caption{\textbf{An example.} }
  \label{fig:illus}
\end{figure}

In our context, as example shown in Figure \ref{fig:illus}, the aim of our proposed attack method is to find a vector of parameters $\theta$ of the laser beam $l$ that makes the resultant image $x_{l_{\theta}}$ being misclassified by the target model $f$. We constrain the width $w$ of laser beam $l$ to satisfy the requirement on the stealthiness of $x_{adv}$ in digital-setting.

This section is organized as follows. We first present the definition of the laser beam. Then we model the physical laser beams with a set of parameters and optimize these parameters to create an adversarial example for the given image $x$.
\subsection{Laser Beam Definition} 
Laser is distinguished from other light sources by its coherence, including temporal coherence and spatial coherence. With high temporal coherence, laser is able to emit light with a very narrow spectrum even in a single color. The spatial coherence enables a laser beam to stay narrow over a long distance. Our proposed attack is based on these unique properties of the laser beam.
We formulate the laser beam $l$ with a set of physical parameters denoted by $\theta$. 
We consider four key parameters to define a laser beam $l$ including wavelength ($\lambda$), layout ($r$, $b$), width ($w$), and intensity $\alpha$. We define each parameter as follows. 
\begin{itemize}
    \item \textbf{Wavelength ($\lambda$).} Wavelength ($\lambda$) denotes the color of the laser beam. We only consider wavelengths in the range of visible light (380 nm to 750 nm). We define a conversion function that converts $\lambda$ to a RGB tuple  \footnote{\url{http://www.noah.org/wiki/Wavelength_to_RGB_in_Python}}. 
    \item \textbf{Layout ($r$, $b$).} We treat the laser beam as a line. We use a tuple ($r$, $b$), which includes an angle ($r$) and intercept ($b$) to determine the beam line. The propagation path of the laser beam can be denoted by $y = \tan(r) \cdot x + b$. We also consider that the laser beam extincts during the transmission. Thus we define an attenuation function with inverse square \footnote{\url{https://en.wikipedia.org/wiki/Inverse-square_law}} to simulate the reduction of luminous intensity of laser beam during transmission.
    \item\textbf{Width ($w$).} The width of the laser beam depends on two factors: the distance between the laser beam and the camera, and the coherence degree of the laser beam. The wider the laser beam is, the more perceptible. We set a threshold on the beam width in digital-setting to avoid over-obstructiveness. 
    \item \textbf{Intensity ($\alpha$).}
    The intensity of the laser beam depends on the power of the laser device. Laser beam with stronger intensity looks brighter. In our context, we use $\alpha$ to denote the intensity of the laser beam $l$. 
\end{itemize}
We define constraint vectors ${\epsilon}_{{min}}$ and ${\epsilon}_{{max}}$ to restrict the range of each parameter in $\theta$. The constraints are adjustable.
We then adopt a simple linear image fusion method to fuse clean image $x$ and laser beam layer $l_{\theta}$.   
\begin{equation}\label{eq:1}
\begin{aligned}
 x_{l_{\theta}} = x + l_{\theta}
\end{aligned}
\end{equation}
where $x_{l_{\theta}}$ represents the image $x$ imposed by a specific laser beam $l_{\theta}$ defined by a vector of parameters $\theta$, then clipped to a valid range.

\subsection{Laser Beam Adversarial Attack} \label{sec:physical-adaptation}
\textbf{Algorithm.}
The focus of the proposed attack is to search for a vector of physical parameters $\theta$ of laser beam $l$ given image $x$, aiming to result in misclassification by $f$. We define a search space $\Theta$, formally, $\Theta=\{\theta | \theta = [\lambda, r,b,w,\alpha], \; {\epsilon}_{min} \leq \theta \leq  {\epsilon}_{max} \}$, where $\epsilon$ is a list of constraints.
In our context we consider the most practical scenario: an attacker cannot attain the knowledge of the target model but only the confidence score $f_{y}(x)$ with given input image $x$ on predicted label $y$. In our proposed $AdvLB$, we exploit by using the confidence score as the adversarial loss. Thus the objective is formulated as minimizing the confidence score on correct label, we construct an adversarial image $x_{l_{\theta}}$ by solving the following objective:
\begin{equation}\label{eq:2}
\begin{aligned}
 \argmin_{\theta} \; f_{y}(x_{l_{\theta}}),\; \text{s.t.}\; \theta \in [{\epsilon}_{min}, {\epsilon}_{max} ]
\end{aligned}
\end{equation}
Inspired by Guo et al.'s work \cite{guo2019simple}, we exploit the confidence scores given by the target model, and propose a greedy method to search for vector $\theta$ to generate adversarial laser beam. In a brief, we repeatedly attempt to update the current vector of parameters $\theta$ with one of stronger adversarial strength instructed by $f_{y}(x_{l_{\theta'}})$, which indicates the confidence of correct class $y$ given laser beam with current parameter $\theta'$. If $f_{y}(x_{l_{\theta'}})$ decreases, we then update the vector of parameters $\theta$ with $\theta'$. We first define the basis for search as set $Q$, which includes a series of predefined candidate vectors $q$. We define a candidate $q$ as follows: the minimal update on $l_{\theta}$ is one unit in either $\lambda$, $r$, $b$, $w$ or $\alpha$. We set the max step numbers with $t_{max}$. During the search, we pick a random vector $q \in Q$ multiplying with step size $s \in S$. If either $f_y(x+l_{\theta + q})$  or $f_y(x+l_{\theta - q})$ decreases, then we update $\theta$ by current $\theta'$. The search ends when the current $x_{l_{\theta}}$ is predicted with a label $ \argmax f(x_{l_{\theta}}) \neq \argmax f(x)$ or the max steps $t_{max}$ is reached. 

However, we found such a greedy search process is prone to getting stuck into local optimum. To this end we further apply a strategy called $k$-random-restart, which introduces more randomness into the search algorithm. Specifically, $k$-random-restart restarts the search process $k$ times. For each time we use $\theta$ with different initializations. We find such a simple strategy greatly improves the effectiveness of the search process. The pseudo-code of $AdvLB$ is shown in Algorithm \ref{alg:alg1}.
\begin{algorithm}[htb]
\SetAlgoLined
\KwInput{Input $x$; Candidate vectors $Q$; Step size $S$; classifier $f$; Max \#steps $t_{max}$; }
\KwOutput{A vector of parameters $\theta$;}
$conf^{\ast} \gets f_y(x)$\;
\For{$i \gets 1\ \mathrm{to} \ k$}{
 Initialization: $\theta \sim  \Theta(\lambda, r, b, w, \alpha)$ \;
 \For {$t \gets 1 \ \mathrm{to} \ t_{max}$}{
 Randomly pick $q \in Q$, $s \in S$\;
 $q \gets q\times s$ \;
$\theta' \gets \theta \pm q$\;
$\theta' \gets \text{clip}({\theta', \epsilon}_{min}, {\epsilon}_{max})$\;
$conf= f_{y}(x_{l_{\theta'}})$\;
 \If{$conf^{\ast} \geq conf$}{
 $\theta \gets \theta^{'}$\;
 $conf^{\ast} \gets conf$\;
 break;
    }
\If{$\argmax f(x_{l_{\theta}}) \neq \argmax f(x) $}{\Return { $\theta$}}
 }
 }
 \caption{Pseudocode of $AdvLB$}
 \label{alg:alg1}
\end{algorithm}


As the algorithm illustrates, the proposed method takes a test image $x$, a set of candidate $Q$, a flexible step size $S$, classifier $f$ and max steps $t_{max}$ as input decided by the attacker. 
Details of the algorithm have been explained above. The algorithm finally returns a successful parameter list $\theta$ of laser beam, which is used for further instructing the deployment of the attack in the physical-world. 

\textbf{Physical Adaptation}
To make adversaries generated by $AdvLB$ physically realized, we adopt two strategies. 1) Physically adapted constraints $\epsilon$. We consider the practical limitations for an attacker to perform the attack. To this end, we denote that $\epsilon$ is physically adapted according to the real-world conditions to perform the attack. 2) Batch of transformed inputs. $AdvLB$ instructs where to perform an effective attack, however, a laser beam with exact layout ($r$, $b$) could be hard to reproduce. Thus we apply $AdvLB$ on a batch of transformed images $X_T = \{x'|x' = T(x) \}$ where $T$ represents a random transformation function including rotation, translation, or addition of noise. With returned batch of $\theta$ on given $x$, we then have an effective range, \eg an effective range of angle $r$, to perform laser attack in the physical-world. 

\section{Evaluation}\label{sec:eval}
\subsection{Experimental Setting}
\begin{figure}[htb]
  \setlength{\abovecaptionskip}{-0.2cm}
 \begin{center}
    \includegraphics[width = 0.8\linewidth]{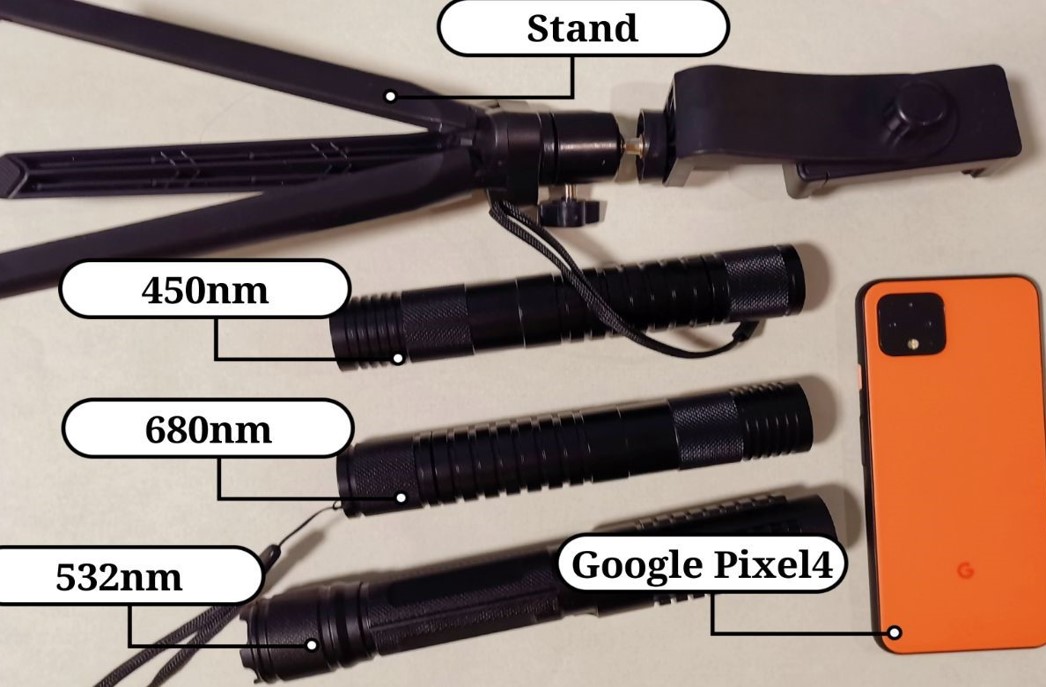}
  \end{center}
  \caption{\textbf{Experiment devices.} }
  \label{fig:devices}
\end{figure}

We test our proposed attacks in both digital- and physical-settings. We use ResNet50 \cite{he2016deep} as the target model for all the experiments. 
We randomly selected 1000 correctly classified images of ImageNet to evaluate proposed $AdvLB$ in digital-setting. For the physical-world experiments, our experimental devices are shown in Figure \ref{fig:devices}. 
In terms of laser beam, we use three small handheld laser pointers (power: 5mW) to generate low-powered laser beams with wavelength 450nm, 532nm, and 680nm respectively. We use a Google Pixel4 smartphone to take photos.
In ablation study, we set a group of experiments to test the impact of different parameters on the adversarial effects of laser beam. The target models adopt a black-box setting that during the attack we only use the prediction scores given by the model. For all the tests we use attack success rate (\%) as the metric to report effectiveness, which is the proportion of successful attacks among the total number of test images examined. 
\begin{figure*}[t]
\setlength{\abovecaptionskip}{-0.2cm}
\setlength{\belowcaptionskip}{-0.2cm}
 \begin{center}
    \includegraphics[width = 0.9 \linewidth]{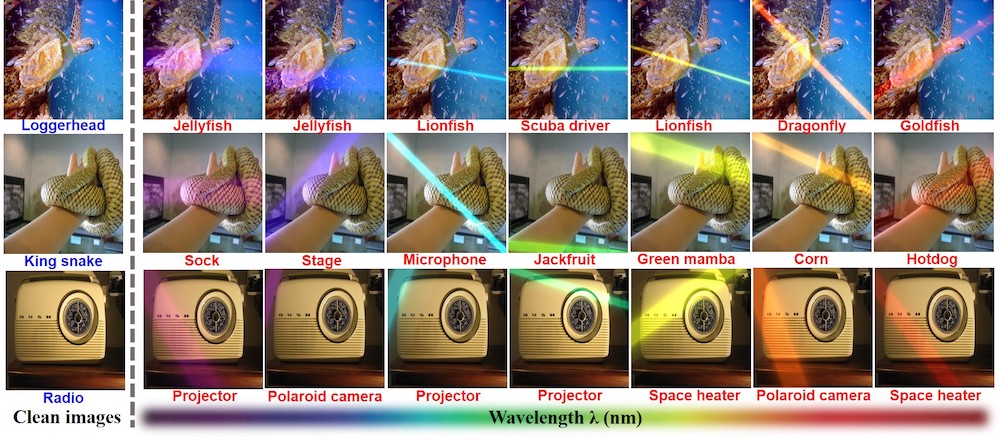}
  \end{center}
  \caption{\textbf{Adversarial examples generated by $AdvLB$.}}
  \label{fig:adv-digital}
\end{figure*}
\subsection{Evaluation of $AdvLB$}
We first evaluate our proposed $AdvLB$ in a digital setting, then we demonstrates how our proposed $AdvLB$ could be a useful tool to explore the real-world adversarial examples resulted from laser beam. 

\textbf{Digital Test}
We apply our proposed attack method on 1000 correctly classified images selected from ImageNet, and craft an adversarial example for each test image with simulated laser beam. The success rate is $95.1\%$ with 834 queries on average of proposed $AdvLB$.  
We also present some attack results generated by $AdvLB$ as shown in Figure \ref{fig:adv-digital}. The first column shows the test images we aim to attack, and each row denotes a series of adversarial examples generated by $AdvLB$. Figure \ref{fig:adv-digital} shows some interesting results. For example, when shining a laser beam with yellow color, the king snake is then misclassified as corn, and there is indeed some similarity between the texture of king snake and corn. Other adversarial examples show similar phenomenon: 'Loggerhead' + Laser beam (blue) $\longrightarrow$ 'Jellyfish', 'Radio' + Laser beam (red) $\longrightarrow$ 'Space heater'. The results show the link among the adversarial class, original class, and the laser beam with a specific wavelength. We will give more discussion in Section \ref{sec:analysis}.

\textbf{Physical Test}
$AdvLB$ aims to be an effective tool to explore real-world threats of laser beam on DNNs. Different from targeted attack in the physical-world, whose evaluation on success is rather intuitive: whether it can fool the DNNs with the class that attacker expects consistently in the physical-world. While untargeted attack is defined as fooling the DNN into any incorrect classes. With amount of uncertainties in the physical-world environment, the misclassification could be caused by natural noise rather than the untargeted attack. Thus evaluation on the effectiveness of untargeted attack in physical-setting requires more careful design of the experiments.

\begin{figure}[htb]
\setlength{\abovecaptionskip}{-0.2cm}
 \begin{center}
    \includegraphics[width = \linewidth]{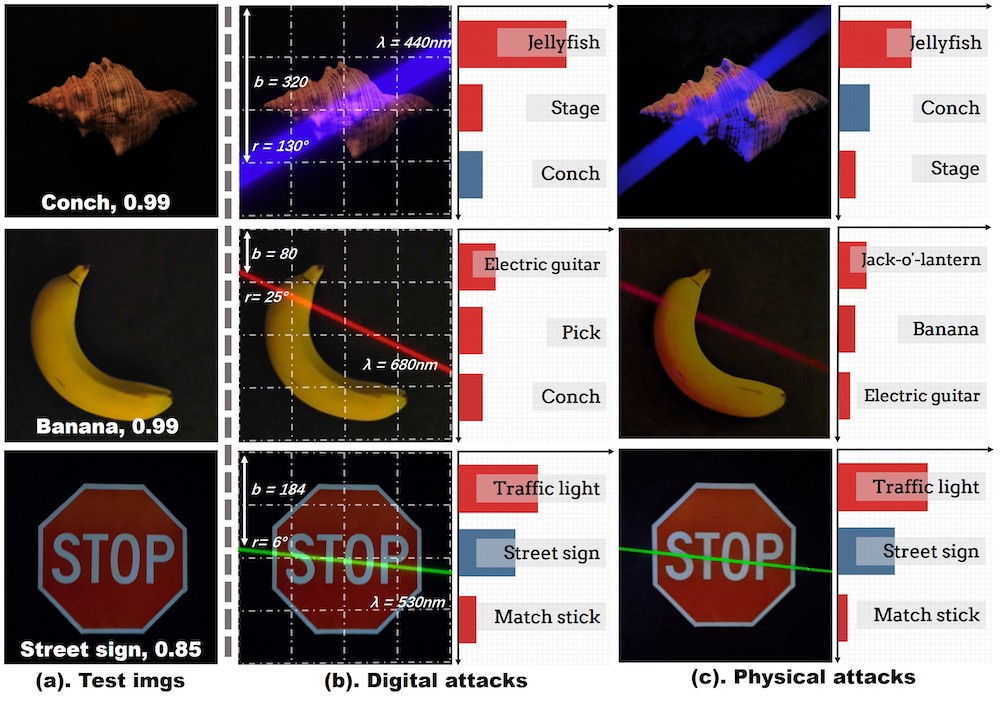}
  \end{center}
  \caption{\textbf{Indoor test.}}
  \label{fig:simulation-test}
\end{figure}

To validate $AdvLB$ can be reproduced by laser beam in the physical-world, we design a strict experimental setting to perform an indoor test. We use three different laser pointers with wavelengths of 450nm, 532nm, 680nm respectively to perform the attack. The target objects include a banana, a conch, and a stop sign. We set the background in black to avoid introducing unnecessary noise into the experiment. For the test, we first capture the target object by cellphone, then we use our proposed $AdvLB$ to find where to perform attack with given test captured image. We then reproduce the attack in such a setting with the returned parameter list $\theta$ by $AdvLB$. We set constraint on parameter $\lambda$ according to the wavelength of used laser pointer. The experimental results are summarized in Figure \ref{fig:simulation-test}.

As Figure \ref{fig:simulation-test} shows, the proposed $AdvLB$ is able to achieve 100\% attack success rate in such a strict experimental setting. Digital attacks by $AdvLB$ are almost consistent with physical-world attacks by reproduced attack by laser pointers, that the top-3 classification results are similar. In which, the laser beam with $\lambda = 680nm$ is harder to capture and the beam is not as coherent as the other two beams, thus the predicted results in digital and physical are slightly different. In summary, our proposed $AdvLB$ can almost reflect the threats of laser beam in the physical-world, thus can be used to explore the potential real-world threats caused by laser beam.

\begin{figure}[hb]
\setlength{\abovecaptionskip}{-0.2cm}
 \begin{center}
    \includegraphics[width = \linewidth]{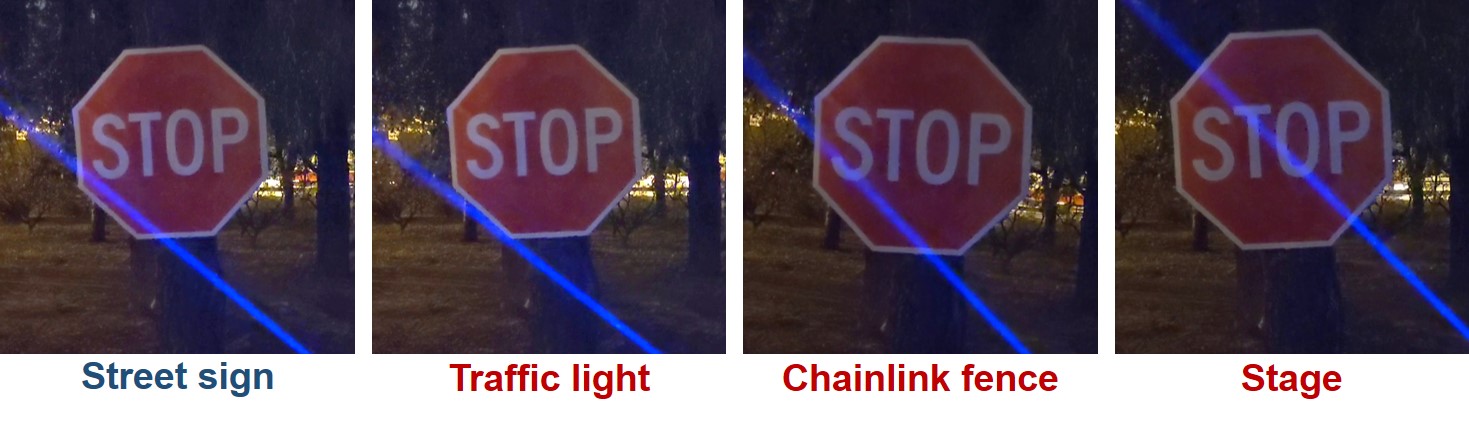}
  \end{center}
  \caption{\textbf{Outdoor test.} }
  \label{fig:video_test}
  \vspace{-0.2cm}
\end{figure}

We further conduct an outdoor test. When a self-driving car approaches the stop sign, even if it fails to recognize the stop sign for merely a short time window, it can lead to a fatal accident. We first apply the method mentioned in Section \ref{sec:physical-adaptation} with given captured stop sign. We then shoot the laser beam from position onto targeted stop sign with given returned effective attack range. Figure \ref{fig:video_test} illustrates some attack results. Overall, there is an attack success rate of 77.43\% of the test.
These results further demonstrate the real-world threats by laser beams. Thus our proposed $AdvLB$ could be a very meaningful tool to explore such threats. Currently, a weakness of our proposed method is that it is still limited in attacking in a dynamic environment, we leave this as future work.

\subsection{Ablation Study}
Here, we conduct a series of experiments on ImageNet to study the adversarial effect of laser beam with different parameters: 1) Wavelength ($\lambda$), 2) Layout ($r$, b), 3) Width ($w$), and another ablation study on how $k$ in $k$-random-restart impacts the attack success rate of $AdvLB$. We acknowledge the intensity $\alpha$ is an important property, especially for improving stealthiness of laser beam. We will do more study on $\alpha$ in the future work. In the following experiments, we fix it as 1.0. Also, for the study on each parameter of the laser beam, we fix the other parameters. 

\textbf{Wavelength ($\lambda$)}  Here, we show how the wavelength $\lambda$ impacts the adversarial effects of laser beam. We test the adversarial effects of laser beam with wavelength ($\lambda$) in the range of visible light (380 nm, 750nm). We fix other parameters as: $r$ = $45^{\circ}$, $b$ = 0, and $w$ = 20. These values are selected based on extensive experiments, which are effective for finding adversarial images. In this ablation, we perform the tests on ResNet50 with  1000 randomly collected images from ImageNet. We combine the images with laser beams using Eq. \eqref{eq:1} and then feed the resultant images to the target model. Table \ref{tab-wave} shows the success rates of simulated laser beams with different wavelength $\lambda$.
\begin{table}[htb]
\caption{\textbf{Ablation of Wavelength ($\lambda$).}}
\label{tab-wave}
\vspace{-0.5cm}
\label{tab-beta}
\begin{center}
 \resizebox{\linewidth}{5mm}{ 
 \begin{tabular}{lccccc}
\hline
\textbf{Wavelength $\lambda$ (nm)} & 380  & 480 & 580 & 680     \\ \hline
\textbf{Suc. rate (\%)} & 34.03 & 48.01 & 58.93 & 44.10         \\ \hline
\end{tabular}}  
\end{center}
\end{table}
\vspace{-0.5cm}

As shown in Table \ref{tab-wave}, the simulated laser beam with $\lambda = 580$ can even achieve 58.9\% success rate. Note that for all these experiments, the simulated laser beams are added on unknown images and then fed to unknown target models directly. The results show that the laser beams have universal adversarial effects on different images.   

\textbf{Width ($w$)}
We then evaluate how the width $w$ impacts the adversarial effect of laser beam. We set the threshold of width as 40, occupied at most 1/10 of the whole image. Again, we fix other parameters of laser beam as constants: $\lambda = 400$, $r = 30$ and $b = 50$. Wider laser beam can improve the success rate from 30.80\% up to 47.69\% with width from 1 to 40. However, even with the smallest $w=1$ (1 pixel width), there is still an impressive success rate (30.80\%). 

\begin{figure}[h]
\setlength{\abovecaptionskip}{-0.2cm}
 \begin{center}
    \includegraphics[width = 0.9 \linewidth]{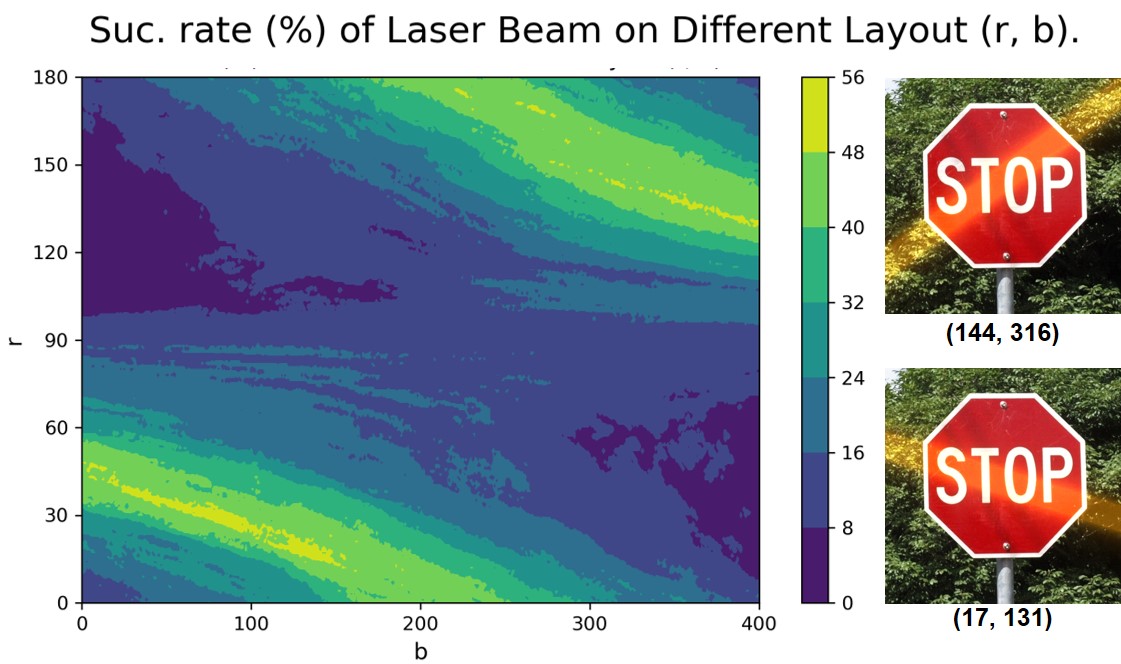}
  \end{center}
  \caption{\textbf{Ablation of Layout ($r$, $b$).}}
  \label{fig:eval-position}
\end{figure}

\textbf{Layout ($r$, $b$)}
We then show how the selection of layout ($r$, $b$) impacts the attack success rates of laser beams. The range of $r$ is [$0^{\circ}$, $180^{\circ}$], and $b$ in range [0, 400]. We fix other physical parameters as constants: $\lambda = 580$, $w$ = 20. For efficiency concern, we only sample 100 correctly classified images from ImageNet, and summarize the results in Figure \ref{fig:eval-position}, where each point denotes the success rate of current layout.
\vspace{-0.2cm}
\vspace{-0.2cm}

In Figure \ref{fig:eval-position}, the left column shows that the attack success rate of laser beam is highly related to its layout. Meanwhile, the right column illustrates two adversarial images (layout with higher attack success rates), which indicate that the center (where the laser beam illuminates) is more likely to create a successful adversarial example.

\textbf{$k$-random-restart ($k$)} 
We choose different $k$ of 1, 20, 50, 100, 200 to run the experiment. The results are shown in Table \ref{tab-k}, the attack success rate is improved gradually with increase of $k$. It suggests that we do find better parameters of laser beam with more random restarts. 
\begin{table}[htb]
\caption{\textbf{Ablation of $k$-random-restart ($k$).}}
\label{tab-width}
\vspace{-0.5cm}
\label{tab-k}
\begin{center}
 \begin{tabular}{lccccc}
\hline
\textbf{Restart num. ($k$)}    & 1      & 50    & 100    & 200    \\ \hline
\textbf{Suc. rate (\%)} & 72.80 &  89.60 & 92.20 & 95.10 \\ \hline
\end{tabular}  
\end{center}
\end{table}
\vspace{-0.5cm}

\section{Discussion}\label{sec:analysis}
\begin{figure}[htb]
\setlength{\abovecaptionskip}{-0.2cm}
 \begin{center}
    \includegraphics[width = 0.9 \linewidth]{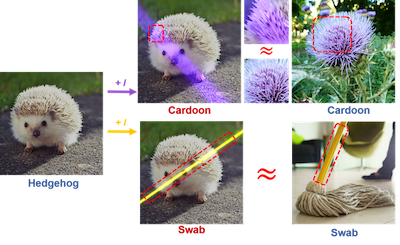}
  \end{center}
  \caption{\textbf{Two types of errors caused by $AdvLB$.} }
  \label{fig:intro-analysis}
\end{figure}

\textbf{Analysis of DNNs' Prediction Errors } 
To better understand the mechanism that the laser beam enables adversarial attacks, we further perform an empirical study with ImageNet to explore the errors caused by the laser beam. Roughly, we find there are two categories of errors as shown in Figure \ref{fig:intro-analysis}.
\vspace{-0.15cm}
The laser beam performs as a kind of perturbation, which either cancels or changes the original feature of a clean image and brings new cues for DNNs. An example is shown in Figure \ref{fig:intro-analysis}, when the laser beam with wavelength 400nm shines on the hedgehog, whose spines combined with the blue color brought by laser beam form a new cue similar to cardoon for the DNN, thus resulting in misclassification. Results shown in Figure \ref{fig:adv-digital} also support for this claim. Also, we find the laser beam itself performs as dominant features for some specific classes (\eg swab shown in Figure \ref{fig:intro-analysis}). We further perform a batch of experiments and then summarize the statistics: which class's percentage rises the most before and after adding the laser beam. We report both top-1 and top-5 rise as shown in Table \ref{tab-error-sum}. For each case, we report both percentages of specific class before and after adding laser beam on ImageNet. 
\begin{table}[htb]
\caption{\textbf{Statistics of Error caused by $AdvLB$.}}
\label{tab-error-sum}
 \resizebox{\linewidth}{9mm}{ 
\begin{tabular}{cllll}
\hline
\textbf{$\lambda$} & \multicolumn{1}{c}{\textbf{Top1 Pred.}} & \multicolumn{1}{c}{\textbf{Percent. (\%)}} & \multicolumn{1}{c}{\textbf{Top5 Pred.}} & \multicolumn{1}{c}{\textbf{Percent. (\%)}} \\ \hline
\textbf{380$\sim$470}           & Feather boa           & 0.10 $\rightarrow$ 2.20       & Feather boa   &  0.32 $\rightarrow$ 8.76  \\
\textbf{470$\sim$560}           & Tennis ball           & \textcolor{red}{0.10 $\rightarrow$ 2.46}      & Spotlight   & 0.64 $\rightarrow$  7.98\\
\textbf{560$\sim$650}           & Rapeseed              & 0.13 $\rightarrow$ 1.94       & Candle       & \textcolor{red}{0.19 $\rightarrow$ 6.21 } \\
\textbf{650$\sim$740}           & Volcano                & 0.11 $\rightarrow$ 2.14      & Gold fish     & 0.26 $\rightarrow$ 6.63 \\ \hline
\end{tabular}}
\end{table}

\begin{figure}[hb]
  \centering
  \begin{subfigure}[b]{\linewidth}
    \includegraphics[width = \linewidth]{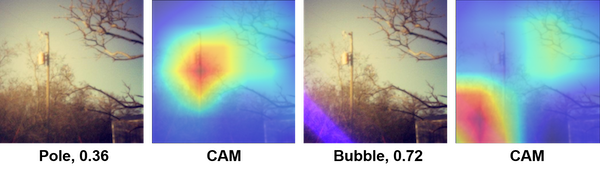}
  \end{subfigure}
  \begin{subfigure}[b]{\linewidth}
    \includegraphics[width = \linewidth]{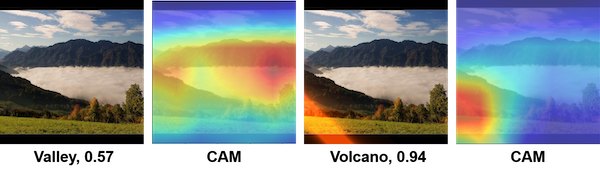}
  \end{subfigure}
  \caption{\textbf{CAM for test images and adversarial images generated by $AdvLB$.}}
  \label{fig:cam}
\end{figure}

As shown in Table \ref{tab-error-sum}, the laser beam with different wavelengths $\lambda$ indeed increases the percentage of some specific classes.
For example, when adding a laser beam with wavelength $\lambda = 380$, the percentage of class "Candle" increases the most from 0.19\% to 6.21\%.
The results imply that the laser beam itself serves as dominant feature for some classes, such as spotlight, volcano, \etc. Thus when adding the laser beam to the clean image, it has the chance that the model is more biased towards the feature brought by the laser beam. We further use the CAM \cite{zhou2016cvpr} to highlight the bias of model when the laser beam is added to the clean image (as shown in Figure \ref{fig:cam}). By adding light beam even on the corner of image, the model is more biased towards "Bubble" and "Volcano", thus give wrong top-1 predictions.
\vspace{-0.15 cm}
\vspace{-0.15 cm}

\textbf{Defense of Adversarial Laser Beam}
Besides revealing the potential threats of $AdvLB$, in this work, we also try to suggest an effective defense for laser beam attack. Similar to adversarial training, we progressively improve the robustness by injecting the laser beam as perturbations into the data for training. We do not find the worst-case perturbations at each training step like adversarial training \cite{madry2017towards}. As seeking worst-case laser beam perturbations by $AdvLB$ needs much more computation cost, it may cause the overall training unaffordable. By contrast, we find that training with randomly added laser beams as augmentation can partly strengthen the model under laser beam attacks, and has no negative impact on the recognition of clean images. 
We use timm\footnote{\url{https://github.com/rwightman/pytorch-image-models}} to train the ResNet50 robust model. The model was optimized on 16 2080Ti GPUs by SGD with a cosine annealing learning rate schedule from 0.1 to 1e-5. We add random laser beams on input images with 50\% probability for data augmentation. 
The other hyperparameters are consistent with the reported settings\footnote{\url{https://github.com/rwightman/pytorch-image-models/blob/master/docs/results.md}}. We summarize the results in Table \ref{tab-digital}. 
\begin{table}[htb]
\caption{\textbf{Comparison of ResNet50 with and without Defense on $AdvLB$.}}
\vspace{-0.35cm}
\label{tab-digital}
\begin{center}
 \begin{tabular}{cccccc}
\hline
\textbf{Models}     & \textbf{Std. acc(\%)} & \textbf{Suc. rate(\%)} & \textbf{Queries} \\  \hline

\textbf{ResNet50$_{ori}$}  & 78.19 & 95.10 & 834                                    \\
\textbf{ResNet50$_{rob}$} & 78.40 \textcolor{red}{($\uparrow$ 0.21)} & 77.20  & 2576                             \\ \hline
\end{tabular}  
\end{center}
\vspace{-0.35cm}
\end{table}
Except for the attack success rate, we adopt another metric named queries. Queries are the number of times that an attacker needs to query the output from target model before searching for best parameters to attack successfully. Our $AdvLB$ only uses 834 queries on average to break through the original ResNet50$_{ori}$ with a high success rate of 95.1\%. In contrast, the defense model ResNet50$_{rob}$ can effectively reduce the attack success rate to 77.2\% based on running 2576 queries, showing a certain degree of defense ability against $AdvLB$. Besides, we found an intriguing phenomenon that the accuracy of the model on clean images does not decrease after adding random laser beam augmentation, instead increases slightly by 0.21\%. We suggest that introducing additional light source as enhancement makes the model more robust to the confusion as analyzed in Section \ref{sec:analysis}, thus increases the generalization ability of DNN.


\section{Conclusion and Future Work}\label{sec:conclusion}
In this paper, we propose $AdvLB$ to utilize the laser beam as adversarial perturbation to generate adversarial examples, which can be applied in both digital- and physical-settings. The proposed attack reveals the existence of an easily implemented real-world threat on DNNs. Some findings resulted from our work open a promising direction for crafting adversarial perturbation by utilizing light (\ie, laser beam) rather than generating perturbation manually. The proposed $AdvLB$ is particularly useful in studying the security threats of vision systems at poor light conditions, that could be a meaningful complementary to current physical-world attacks.

In the future, we will improve our proposed $AdvLB$ to be more adapted to dynamic environment. In addition, we will consider the parameter of light intensity into optimization to create a more stealthy adversarial example with simulated laser beam. We will also explore the possibility of using other light pattern (\eg spot light) and light source (\eg natural light) to craft adversarial attacks. Furthermore, we will apply $AdvLB$ on other computer vision tasks including object detection and segmentation. Moreover, effective defense strategies against such attacks will be another crucial and promising direction.

\bibliographystyle{plain}
\bibliography{adv_light}

\end{document}